# Benchmark Evaluation of Image Fusion algorithms for Smartphone Camera Capture

Lucas Nedel Kirsten
Lenovo Research
Motorola Mobility Comércio de Produtos Eletrônicos Ltda
Jaguariúna, SP 13918-900, Brazil
lucask@motorola.com

**Abstract**— This paper investigates the trade-off between computational resource utilization and image quality in the context of image fusion techniques for smartphone camera capture. The study explores various combinations of fusion methods, fusion weights, number of frames, and stacking (a.k.a. merging) techniques using a proprietary dataset of images captured with Motorola smartphones. The objective was to identify optimal configurations that balance computational efficiency with image quality. Our results indicate that multi-scale methods and their single-scale fusion counterparts return similar image quality measures and runtime, but single-scale ones have lower memory usage. Furthermore, we identified that fusion methods operating in the YUV color space yield better performance in terms of image quality, resource utilization, and runtime. The study also shows that fusion weights have an overall small impact on image quality, runtime, and memory. Moreover, our results reveal that increasing the number of highly exposed input frames does not necessarily improve image quality and comes with a corresponding increase in computational resources usage and runtime; and that stacking methods, although reducing memory usage, may compromise image quality. Finally, our work underscores the importance of thoughtful configuration selection for image fusion techniques in constrained environments and offers insights for future image fusion method development, particularly in the realm of smartphone applications.

**Keywords**- Image Fusion; Smartphone image capture.

## I. INTRODUCTION

Photographs captured on environments with highly imbalanced lighting usually result in a poor-quality picture mainly due to the presence of under- and over-saturated regions. This is caused by the limited dynamic range of digital cameras, i.e. the ratio of the highest brightness to the lowest brightness in a scene [1]. For most cameras, this dynamic range is on the order of 2 ~ 3, while real scenes can display orders of 10 or higher [2]. Acquiring multiple exposure frames of the same scene allows capturing the high dynamic range; however, upon display, the intensities need to be remapped to match the device's range through tone mapping [3] or with exposure fusion methods [4; 5; 6].

Exposure fusion methods rely on computing perceptual quality measures (a.k.a., weight maps) for each pixel in the multi-exposure low-dynamic-range (LDR) sequence of frames, and then selecting the "good" pixels using some weighted merge-like operation (e.g., mean or median) [1; 5] – suggesting that the choice of the weight maps must have a high impact on the quality of the resulting image. Exposure methods have several advantages compared to other methods (e.g., tone mapping), such as a simplified pipeline (since no in-between HDR image needs to be computed [7]), it does not require camera calibration, and can even use images captured with external assistance (e.g., with flash) to improve the results [1].

Most popular fusion methods compute the final "merged" image using Laplacian pyramids (a.k.a., multi-scale image fusion methods), in which a sequence of sub-images is generated to produce more natural transitions on the edges of the merged image [1; 2]. However, the multiple operations required for computing the pyramid sub-images can introduce large computational overhead depending on the number and the size of the input images, especially in hardware-constrained applications (e.g., smartphones). One workaround is to combine images (a.k.a., image stacking or image merging) with close Exposure Value (EV) using some reduction function (e.g., mean, median), to diminish the number of input images [4; 5]. Single-scale image fusion methods have also been proposed to alleviate the pyramid computational overhead by approximating the overall impact of the Laplacian pyramids [8]. However, as observed in [2; 5], single-scale methods usually produce worse image quality due to the huge gray difference of the images to be fused, creating obvious seams. Yet, is the overhead introduced by multi-scale methods justified in the quality of the merged image when compared to single-scale methods? What is the individual impact of the weight maps on the final image? What is the benchmark between the used computational resources and image quality related to using more input frames? And do stacking methods collaborate to reduce the required computation resources without losing image quality?

To answer these questions, in this work, we investigate the individual and joined impacts on the used computational resources, runtime, and image quality related to the variables: fusion method, weights of the fusion method, stacking method, and used number of frames. The main objective of this work is to determine the benchmark among these variables and to establish the best combinations to be employed in hardware-constrained environments, such as smartphones. We evaluated multiple combinations of these variables on a proprietary dataset composed of Motorola smartphone pictures collected in diverse environments and under various lighting conditions. Our

**3911**

evaluation protocol used standard literature metrics for image quality and correlated them with the used computational resource and runtime for each variable combination. Moreover, although the current literature has mainly focused on deep learning-based approaches [4; 6], we emphasize that our findings can be directly used for such methods since some of the evaluated variables (such as the number of frames to be used, and the stacking method) are also important hyperparameters employed in such learning algorithms.

## II. RELATED WORK

Multi-scale image fusion methods refer to approaches that employ a sequence of Laplacian pyramids to decompose the details of the input frames (and weight maps) into "sub images" (i.e., downsized versions of the input frames), to eliminate exposure differences and make the local transitions more natural [2]. The pioneer Merge Mertens [1] algorithm is based on computing three quality measures related to the image saturation, contrast, and exposure levels of the input frames, and then employing the Laplacian pyramids sequence on both input frames and weight maps. The resulting sub images and maps are then merged as in a weighted average fashion.

The Merge Mertens algorithm needs to operate in the RGB color channels, which can introduce large overhead, especially for hardware-constrained applications (e.g., smartphones), due to the multiple operations required for computing the pyramid sub-images on all the 3 input channels of each frame. To improve Merge Mertens performance, Liu et al. [2] proposed the Fast YUV method, based on using the YUV color space. The authors propose to compute the pyramid sub-images using only the Y channel (hence reducing the pyramid computational efforts by 1/3 compared to the standard Merge Mertens), using 2 image quality measures (Merge Mertens employs 3), and fusing the UV channels using a maximum reduction operation over the frames sequence.

Both Merge Mertens and Fast YUV depend on computing multiple sub-images through a sequence of Laplacian pyramids. Depending on the size and number of sequence frames, this process can demand a huge amount of computational resources, impacting both memory usage and runtime. An easy assumption would be to make all computations in the original image resolution. However, as noted in [2], due to the huge gray difference of the images to be fused, it will usually lead to unnatural transitions of the edges, creating obvious seams. Nevertheless, Ancuti et al. [8] proposed a single-scale image fusion (namely, SSF) method based on approximating the operations in the pyramid step of the standard Merge Mertens algorithm. Although their work was not capable of generally overcoming the reference multi-scale fusion method, it highly improved the usual edge transitions issues referred to in [2], and considerably reduced the computation resources requirements.

More recently, methods based on deep learning have also been explored for image fusion [4; 6; 9]. The main advantage of such techniques is their capability to learn from examples, so there is no necessity to manually define the image quality measures for the weight maps. However, such methods are designed to work with specific input setups (e.g., they require the same number of frames with defined EV to work properly), which is usually not the case on real applications. For instance, the frames' EV can depend on the scene detected illuminance, and the number of frames can depend on predefined computational restrictions for capture, or on the output of some "frame selector" algorithm (e.g., one that eliminates poor quality frames). Moreover, in the scope of this work, evaluating learning-based methods would require a full retraining step for each considered variable combination to establish a fair baseline comparison, making it timely unfeasible (see Section III-C for more details). Nevertheless, our findings can be directly explored to define better setups for such methods (e.g., number of frames, stacking method to be used) as we discuss in more detail in Sections IV and V.

## III. EXPERIMENTAL SETUP

In this work, we investigate the benchmark between the computational resource (i.e., runtime and memory usage) and image quality measures related to the combination of the following variables on fusing LDR images: fusion method, weights of the fusion method, used number of frames, and stacking method. We employed a proprietary dataset composed of Motorola smartphone pictures collected using the front camera in diverse environments and under various lighting conditions. Nevertheless, we emphasize that the usual complete process of mobile image capture relies on more steps than the ones being considered in this work, such as frame selection for removing low-quality frames, frame alignment, and image enhancement (e.g., color correction, and contrast improvement). However, considering all these steps as variables would be unfeasible due to the huge number of possible combinations to be evaluated. Hence, we limit our experiments to the aforementioned variables, which have a higher impact on the fusion process itself [4; 5; 6]. We proceed to provide details regarding our complete experimental setup.

*A.    Dataset*

We employed a proprietary dataset composed of 480 Motorola smartphone captured images using its 8-megapixel front camera. The images were collected in several different indoor and outdoor scenes with environment illuminance ranging from 0,15 lux up to 20,2 lux. Six frames were captured for each scene with EV values: -24, 0, 1, 2, 3, and 4. The order in which the frames were captured is from the lowest (EV -24) to the highest (EV 4). The ground-truth (GT) images were manually generated using image processing software to obtain the "best quality" merged image. The 6 captured frames plus a very long exposed frame (which is not available during the usual device's camera capture, and was used for color improvement and correction) were accessible to the annotators. However, image quality is generally a subjective task [10] (i.e., different annotators will usually provide distinct GT images for the same set of frames), and hence, the following protocol was established to standardize the development of the GT images:

•    Regarding brightness and color, GT should be comparable to EV 3 frame;
•    Regarding clarity (details), GT should be comparable to EV 0 and EV 3 frames;
•    Artifacts, such as noise, must be removed in the GT;
•    Over- and underexposure areas must be corrected in the GT.

We provide examples of the dataset in Fig 1.

**3912**

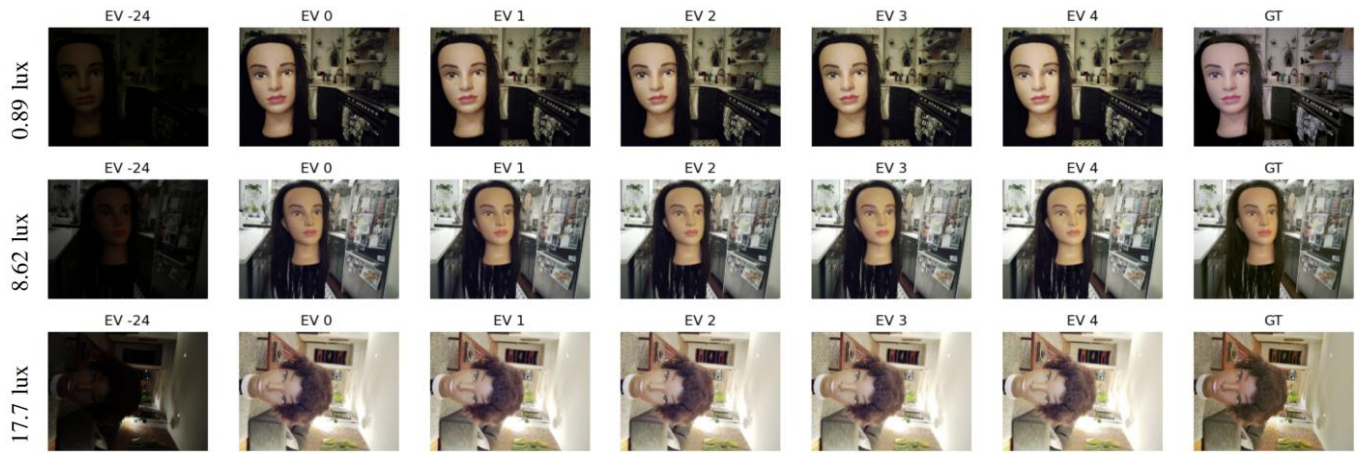

Figure 1. Example of image frames for three scenes with different illuminance levels (value in the left) from the employed dataset.

## B. Variable Combinations

Each of our dataset images was evaluated using all possible combinations for each of the following variables: fusion method, weights of the fusion method, used number of frames, and stacking method (54,850 total tested combinations). Tab I summarizes the space of values for each of the considered variables, where one "variable combination" refers to choosing one-row value for each column of this table. We proceed to provide details regarding each of the considered variables.

TABLE I. VARIABLES SPACE OF VALUES SUMMARY. EACH VARIABLE COMBINATION REFERS TO CHOOSING ONE-ROW VALUE FOR EACH COLUMN. $W_C$ REFERS TO CONTRAST WEIGHT, $W_E$, TO EXPOSURE WEIGHT, AND $W_S$ TO SATURATION WEIGHT.

| Fusion Method | Fusion Weights | EV≥0 Frames | Stacking Method |
|---|---|---|---|
| Merge Mertens | Wc, W_E, W_S* | 1 | Mean† |
| Fast YUV | Wc, W_E | 2 | Median† |
| SSF RGB | Wc, W_S* | 3 | None |
| SSF YUV | W_S*, W_E | 4 | |
| | | 5 | |

*Only applicable to Merge Mertens and SSF RGB. †Only applicable when using more than 1 EV≥0 frame.

### 1) Fusion Method

We implemented the Merge Mertens [1], Fast YUV [2], and the Single-Scale Fusion [8] (namely, SSF RGB) methods using their original proposed implementation. We also implemented an adapted version of the SSF RGB to work on YUV images (namely, SSF YUV), adapting the SSF RGB [8] w.r.t. Fast YUV [2] with the following modifications: we employ the single Laplacian step only on the Y channel; we compute 2 weight maps equal to those of Fast YUV; and the UV channels are fused using the maximum per-frame function. All methods were developed using Python 3 [11] programming language with OpenCV [12] library, and are available at: https://github.com/LucasKirsten/Benchmark-Image-Fusion.

### 2) Fusion Weights

In all fusion methods, the final weight map is computed following:

$$W = \prod_{i=1}^{M}(W_i)^{k_i}, \quad (1)$$

where $W_i \in \mathbb{R}^{W \times H \times N}$ is the $i$-th weight map, $k_i$ is an associated exponential factor (usually ranging from 0 to 1), $N$ is the number of input frames, and $M$ is the number of computed weights: $M = 3$ ($W_C$: contrast, $W_S$: saturation, and $W_E$: exposure) for the RGB-based fusion methods (Merge Mertens and SSF RGB), and $M = 2$ ($W_C$ and $W_E$) for the YUV-based ones (Fast YUV and SSF YUV). Note that, if $k_j = 0$, the corresponding weight map will not have any effect during the fusion process, since $(W_j)^0 = \mathbf{1}$. In the fusion method implementations, we explored this fact to individually evaluate the influence of each proposed weight map. First, we computed the fused image using all the method proposed weights (equivalent to setting all $k = 1$), and then we excluded the computation of one weight map (equivalent to setting one $k = 0$ but without the computational cost of first computing the weight map and then raising it to 0). Specifically, for the RGB methods we used 4 weight combinations (1 using all weights, and the other 3 excluding one weight), and for the YUV methods we used 3 weight combinations (1 using all weights, and the other 2 excluding one weight).

### 3) Number of Frames

We used the single EV negative (EV -24) frame on all tested combinations (to correct overexposure areas), and iterated on the number of EV-positive (EV≥0) ones. Hence, for each combination, we used 1 EV negative + $N$ EV-positive frames, with $N$ ranging from 1 to 5. The EV-positive frames order of choice followed their ascending value (i.e., 0, 1, 2, 3, and 4), which is the order the frames were acquired in the smartphones.

### 4) Stacking Method

We evaluated the usage of the Mean and Median stacking functions on the *EV-positive frames*. Both methods are applied in the form:

$$S = F(\cup_{i=1}^{N} I_i), \quad (2)$$

where $S$ is the stacked image, $I_i \in \mathbb{R}^{W \times H \times C}$ is the i-th image to be stacked, $F: \mathbb{R}^{N \times W \times H \times C} \to \mathbb{R}^{W \times H \times C}$ is the stacking function, and N is the number of frames to be stacked. We also evaluated the effects of not using any stacking method. Hence, in total, three combinations were evaluated: Mean, Median, and "None"

**3913**

(referring to not using any stacking method on the EV-positive frames).

*C.    Evaluation Protocol*

For evaluating the quality of the fused images, we employed three standard literature metrics, namely: Multi-scale Structural Similarity Index (MS-SSIM) [13], Peak signal-to-noise ratio (PSNR), and Erreur Relative Globale Adimensionnelle de Synthese (ERGAS) [14]; and one metric based on deep learn-` ing: Learned Perceptual Image Patch Similarity (LPIPS) [10]. MS-SSIM is a perception-based metric that considers image degradation as the perceived change in structural information, while also incorporating important perceptual aspects, such as luminance, image distortion, and the combination of contrast distortion [5]. PSNR1 is used to compute the ratio of peak power and noise value power. ERGAS is used to quantify the image quality from the fusion of high spatial resolution images. LPIPS employs a large-scale, highly varied, perceptual similarity dataset to fine-tune deep learning models for the image quality assessment task. The image quality experiments were conducted in an AWS 6a.32xlarge instance4 with 3rd generation AMD EPYC processor, 128 CPUs and 256 GB of memory. We leverage the high number of CPUs to parallelize the variable combination computation for each tested image. Nevertheless, the whole computation took more than one week to be completed. In addition, we manually analyzed the images to confirm that the metrics agreed with the visual results.

The runtime and memory experiments used a Motorola smartphone with 4 GB of RAM, four 2.4 GHz Kryo 265 Gold and four 1.9 GHz Kryo 265 Silver processors, Snapdragon 680 4G Qualcomm SM6225 chipset, and Adreno 610 GPU. The algorithms for each variable combination were converted to the TensorFlow Lite format (to run on the devices), and the measurements were performed with the TensorFlow Lite benchmark tool with 10 simulated runs using GPU.

## IV.  RESULTS AND DISCUSSION

Tab IV (at the end of the paper) presents the grouped results for our experiments involving all the variable combinations. Notably, the best performing configuration varies across variable choices, with some consistencies observed in some optimal parameter combinations, as we proceed to discuss. We recall that the usual complete process of mobile image capture relies on more steps than the ones being considered in this work, and so this may be the main cause of some low image quality metrics values. To better interpret the individual variable impacts, we also present the grouped results for the tested fusion methods and weights, merging all stacking methods and number of EV-positive frames results altogether (i.e., for a stacking and frame variation agnostic evaluation) in Tab II. Similarly, in Tab III, we show the grouped results related to stacking methods and numbers of EV-positive frames, merging all fusion methods and weights combinations results altogether (i.e., for a fusion agnostic evaluation).

In regard to the fusion methods and weights, observe that combinations involving Exposure and Contrast weights tend to yield the highest MS-SSIM and PSNR, and the lowest ERGAS values, indicating superior image quality and spectral distortion reduction. However, it is worth noting that all quality measures produce very similar results, implying that these variables have a minimal effect on the final image quality. Related to computational resources, we can note that methods that operate on the YUV color space are faster and more efficient than the ones that operate on RGB. The primary explanation appears to be related to the predominance of operations made exclusively within the Y channel, as opposed to RGB methods which require operations across all three channels. Furthermore, see that Fast YUV had the faster runtime among all methods (including the single-scale ones), and the memory usage was very similar to its single-scale counterpart, SSF YUV. This could be attributed to the fact that, although SSF YUV eliminates the multiple pyramid step, it introduces additional dot product operations, which appear to be more resource-intensive for the tested smartphone hardware.

TABLE II.    GROUPED RESULTS FOR THE FUSION METHODS AND WEIGHTS REPORTING THE MEAN VALUE FOR EACH METRIC. THE TIME COLUMN IS IN SECONDS, AND THE MEMORY IN MEGA-BYTES (MB). BEST RESULTS ARE MARKED IN **BOLD**, WHEREAS WORST RESULTS ARE IN UNDERLINE. WC REFERS TO CONTRAST WEIGHT, WE TO EXPOSURE WEIGHT, AND WS TO SATURATION WEIGHT.

| Fusion Method | Fusion Weights | MS-SSIM | PSNR | LPIPS | ERGAS | Time | Memory |
|---|---|---|---|---|---|---|---|
| Fast YUV | WE | 0.48 | 15.48 | 0.62 | 11.75 | **0.94** | 475.97 |
| | Wc | 0.48 | 15.48 | 0.62 | 11.75 | **0.94** | 473.97 |
| | Wc, WE | 0.48 | 15.48 | 0.62 | 11.75 | 0.95 | 474.10 |
| Mertens | Wc, WE | 0.48 | **15.51** | 0.62 | **11.71** | 2.85 | 1139.73 |
| | Ws, Wc | 0.48 | 15.48 | 0.62 | 11.75 | 3.38 | 1011.64 |
| | Ws, WE | 0.48 | 15.48 | 0.62 | 11.75 | 2.48 | 1800.26 |
| | Ws, Wc, WE | 0.48 | 15.48 | 0.62 | 11.75 | 3.52 | 1201.87 |
| SSF RGB | Wc, WE | 0.48 | 15.48 | 0.62 | 11.75 | 2.60 | 845.17 |
| | Ws, Wc | 0.48 | 15.48 | 0.62 | 11.75 | 3.14 | 719.37 |
| | Ws, WE | 0.48 | 15.48 | 0.62 | 11.75 | 2.44 | 1122.45 |
| | Ws, Wc, WE | 0.48 | 15.48 | 0.62 | 11.75 | 3.29 | 919.19 |
| SSF YUV | WE | 0.48 | 15.48 | 0.62 | 11.75 | 0.96 | 419.37 |
| | Wc | 0.48 | 15.48 | 0.62 | 11.75 | 0.96 | **410.16** |
| | Wc, WE | 0.48 | 15.48 | 0.62 | 11.75 | 0.97 | 415.38 |

Regarding stacking methods, we can observe that, across all frame counts, the None method (i.e., not using any stacking method) consistently outperforms the others in terms of image quality measures, while also demonstrating the lowest processing time when using 3 EV-positive frames or less. Overall, the None stacking method proves to be a robust choice, offering high-quality results with efficient processing across varying numbers of frames. Nevertheless, note that None stacking has the highest memory usage among the other stacking methods (1.9 times more in the worst case, and 1.2 in the best for the same number of frames). These findings are highly significant as they illustrate that opting not to utilize any stacking method consistently yields superior image quality and reduces processing time for image fusion. This conclusion is not immediately intuitive, as one might anticipate that employing some stacking method would at least decrease runtime, given that fewer frames would be supplied to the fusion algorithm.

**3914**

Regarding solely the number of used frames, note that, as expected, using more frames usually requires more processing time and memory, but it doesn't necessarily translate into better image quality. Specifically, note that, when using 3 frames and None stacking, we have the best MS-SSIM score, and with Median we have the best LPIPS score. As previously mentioned, incorporating additional EV-positive frames during fusion involves feeding images with longer exposure. Hence, for this use-case scenario, we observe that using frames with EV higher than 2 doesn't necessarily lead to an improvement in the fused image quality.

TABLE III. GROUPED RESULTS FOR THE TESTED NUMBER OF EV-POSITIVE FRAMES RELATED TO THE EMPLOYED STACKING METHOD REPORTING THE MEAN VALUE FOR EACH METRIC. BEST RESULTS ARE MARKED IN **BOLD**, WHEREAS WORST RESULTS ARE IN <u>UNDERLINE</u>. WC REFERS TO CONTRAST WEIGHT, WE TO EXPOSURE WEIGHT, AND WS TO SATURATION WEIGHT.

| EV≥0 Frames | Stacking Method | MS-SSIM | PSNR | LPIPS | ERGAS | Time | Memory |
|---|---|---|---|---|---|---|---|
| 1 | None | <u>0.45</u> | <u>14.16</u> | 0.63 | <u>13.38</u> | **1.19** | 657.48 |
| 2 | Mean | 0.46 | 14.35 | 0.62 | 13.09 | 1.88 | **651.01** |
| 2 | Median | 0.46 | 14.35 | 0.62 | 13.09 | 2.01 | 704.30 |
| 2 | None | 0.50 | 17.14 | 0.62 | 9.69 | 1.59 | 791.54 |
| 3 | Mean | 0.47 | 14.55 | 0.62 | 12.80 | 2.04 | 662.49 |
| 3 | Median | 0.47 | 14.58 | **0.61** | 12.76 | 2.06 | 740.85 |
| 3 | None | **0.52** | 18.88 | 0.62 | 8.00 | 1.91 | 871.17 |
| 4 | Mean | 0.46 | 14.43 | 0.64 | 12.05 | 2.24 | 673.79 |
| 4 | Median | 0.46 | 14.46 | 0.64 | 12.02 | 2.31 | 779.29 |
| 4 | None | 0.51 | 20.16 | 0.67 | 6.69 | 2.45 | 1250.73 |
| 5 | Mean | 0.47 | 14.63 | 0.64 | 11.78 | 2.38 | 684.95 |
| 5 | Median | 0.46 | 14.68 | 0.64 | 11.72 | 2.44 | 818.24 |
| 5 | None | 0.51 | **20.77** | <u>0.68</u> | **6.32** | <u>2.83</u> | <u>1326.46</u> |

Overall, these findings underscore the importance of carefully selecting the setup for image fusion based on the desired outcome metric and computational constraints. Our visual inspection corroborates with these findings, as illustrated in Figs 2, 3 and 4. Specifically, note in Fig 2 that increasing the number of EV-positive frames above 3 (EV values higher than 2) does not seem to have major effects on improving image quality. Moreover, regarding stacking methods, Fig 3 shows that None stacking produces brighter (clearer) images compared to Mean and Median. Finally, regarding the fusion method and weights, Fig 4 demonstrates that the tested methods and their weight variation produce similar results.

## V. CONCLUSIONS

In this work, we delve into examining the trade-off between computational resources and the quality of images generated by employing different fusion methods, fusion weights, used number of frames, and stacking techniques. Our study used a proprietary dataset comprising images taken with Motorola smartphones' front cameras across different environmental settings and lighting conditions. Our goal was to determine the variable combinations that produce the best image quality related to its runtime and resource allocation. In regard to using multi- or single-scale methods, the literature often highlights that, although single-scale ones should run faster, they usually do not provide good image quality [1; 2; 5]. However, our work shows that the multi-scale Fast YUV [2] had the faster runtime, and the second lowest memory usage among all tested methods. Moreover, both versions of the single-scale SSF [8] (RGB and YUV) had similar image quality results compared to the tested multi-scale methods (Fast YUV and Mertens [1]). Our research also uncovered that methods operating in the YUV color space exhibit superior benchmark performance compared to RGB color-based ones. Specifically, they produce similar image quality results with faster runtime and lower memory usage. This discovery is significant as it suggests that the advancement of new learning-based methods could benefit from utilizing YUV color space images, with a focus on operating solely on the Y color channel to conserve computational resources. Additionally, our findings demonstrate minimal impacts associated with the choice of fusion weights proposed in the literature on final image quality. This further aligns with the recent trend in deep learning methods [4; 6], which aims to learn weight maps based on example data rather than relying on hand-coded ones.

Regarding the number of used frames, we showed that feeding more input frames for the fusion method not necessarily will improve the final image quality. More specifically, our experimental setup demonstrated that using frames with EV value superior to 2 (in our specific case, using 3 frames) usually will not improve the final image quality, but as expected, it will require more computation resources, increasing runtime and memory usage. To address this limitation, we also investigated stacking methods to decrease the number of EV-positive input frames fed into the fusion method. However, our experiments revealed that the tested methods (Mean and Median) resulted in lower image quality compared to not using any stacking (None method), despite exhibiting similar runtime. Nevertheless, stacking methods proved to significantly reduce memory usage, particularly when more frames were employed. Regarding the recent trend in deep learning methods, this finding suggests that the architecture of these models could be designed to accommodate many frames and then employ some learning-based stacking method (e.g., using convolutional layers with a reduced number of neurons compared to the input frames) to mitigate the need for extensive memory resources.

Finally, our results underscore the importance of carefully configuring the image fusion setup based on both the target image quality metrics and the computational limitations of the system. Moreover, they offer valuable insights for the advancement of new fusion techniques. In future works, we aim to leverage these discoveries to devise more efficient methods for smartphone image fusion.

## REFERENCES

[1] T. Mertens, J. Kautz, and F. Van Reeth, "Exposure fusion," in *15th Pacific Conference on Computer Graphics and Applications (PG'07)*. IEEE, 2007, pp. 382–390.

[2] Y. Liu, D. Wang, J. Zhang, and X. Chen, "A fast fusion method for multi-exposure image in yuv color space," in *2018 IEEE 3rd Advanced Information Technology, Electronic and Automation Control Conference (IAEAC)*, 2018, pp. 1685–1689.

**3915**

[3] E. Reinhard, W. Heidrich, P. Debevec, S. Pattanaik, G. Ward, and K. Myszkowski, *High dynamic range imaging: acquisition, display, and image-based lighting*. Morgan Kaufmann, 2010.

[4] F. Xu, J. Liu, Y. Song, H. Sun, and X. Wang, "Multiexposure image fusion techniques: A comprehensive review," *Remote Sensing*, vol. 14, no. 3, p. 771, 2022.

[5] H. Kaur, D. Koundal, and V. Kadyan, "Image fusion techniques: a survey," *Archives of computational methods in Engineering*, vol. 28, pp. 4425–4447, 2021.

[6] Y. Liu, X. Chen, Z. Wang, Z. J. Wang, R. K. Ward, and X. Wang, "Deep learning for pixel-level image fusion: Recent advances and future prospects," *Information Fusion*, vol. 42, pp. 158–173, 2018.

[7] R. Fattal, D. Lischinski, and M. Werman, "Gradient domain high dynamic range compression," in *Seminal Graphics Papers: Pushing the Boundaries, Volume 2*, 2023, pp. 671–678.

[8] C. O. Ancuti, C. Ancuti, C. De Vleeschouwer, and A. C. Bovik, "Single-scale fusion: An effective approach to merging images," *IEEE Transactions on Image Processing*, vol. 26, no. 1, pp. 65–78, 2017.

[9] Q. Yan, T. Hu, Y. Sun, H. Tang, Y. Zhu, W. Dong, L. Van Gool, and Y. Zhang, "Towards high-quality hdr deghosting with conditional diffusion models," *IEEE Transactions on Circuits and Systems for Video Technology*, pp. 1–1, 2023.

[10] R. Zhang, P. Isola, A. A. Efros, E. Shechtman, and O. Wang, "The unreasonable effectiveness of deep features as a perceptual metric," in *CVPR*, 2018.

[11] G. Van Rossum and F. L. Drake Jr, *Python reference manual*. Centrum voor Wiskunde en Informatica Amsterdam, 1995.

[12] G. Bradski, "The OpenCV Library," *Dr. Dobb's Journal of Software Tools*, 2000.

[13] Z. Wang, E. Simoncelli, and A. Bovik, "Multiscale structural similarity for image quality assessment," in *The Thrity-Seventh Asilomar Conference on Signals, Systems Computers, 2003*, vol. 2, 2003, pp. 1398–1402 Vol.2.

[14] L. Wald, "Quality of high resolution synthesised images: Is there a simple criterion?" in *Third conference" Fusion of Earth data: merging point measurements, raster maps and remotely sensed images"*. SEE/URISCA, 2000, pp. 99–103.

[15] S. van der Walt, J. L. Schonberger, J. Nunez-Iglesias,¨ F. Boulogne, J. D. Warner, N. Yager, E. Gouillart, T. Yu, and the scikit-image contributors, "scikit-image: image processing in Python," *PeerJ*, vol. 2, p. e453, 6 2014.

TABLE IV. COMPLETE GROUPED RESULTS FOR THE TESTED VARIABLE COMBINATIONS REPORTING THE MEAN VALUE FOR EACH METRIC (SPLIT INTO THREE COLUMNS). THE TIME COLUMN IS IN SECONDS, AND THE MEMORY IS IN MEGA-BYTES (MB). BEST RESULTS ARE MARKED IN **BOLD**, WHEREAS WORST RESULTS ARE IN UNDERLINE. WC REFERS TO CONTRAST WEIGHT, WE TO EXPOSURE WEIGHT, AND WS TO SATURATION WEIGHT.

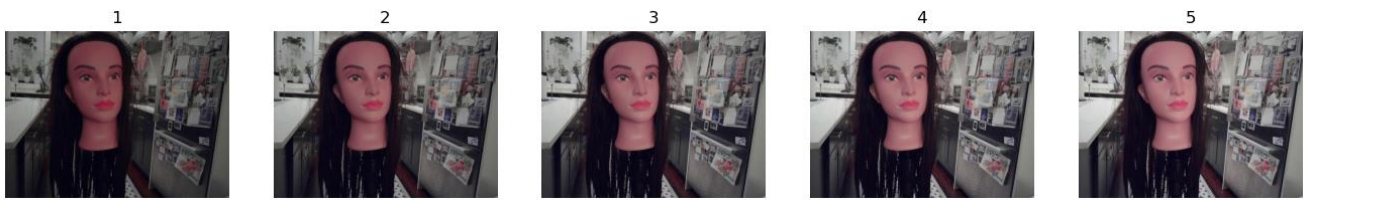

Figure 1. Results for variable number of EV-positive frames. Fixed stacking method to None, and fusion method to Fast YUV with $W_C$ and $W_E$ weights.

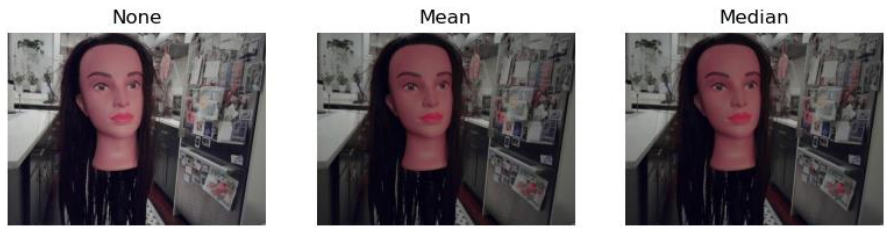

Figure 2. Results for variable stacking method. Fixed number of EV-positive frames to 3, and fusion method to Fast YUV with $W_C$ and $W_E$ weights.

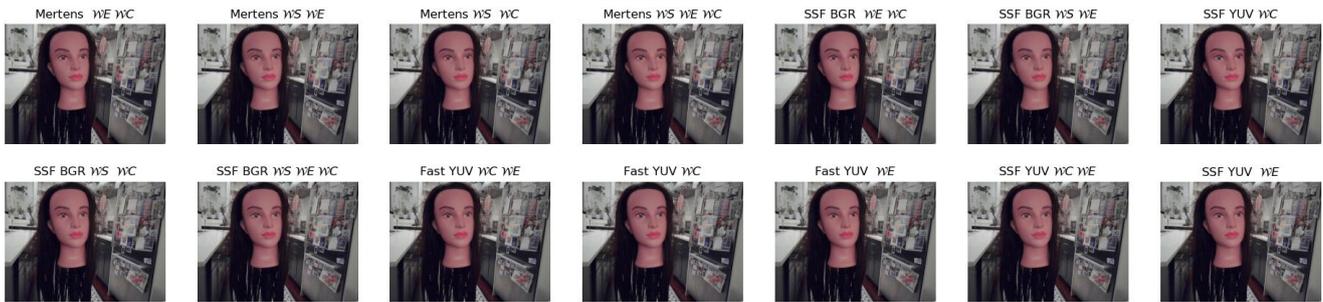

Figure 3. Results for variable fusion method and weights. Fixed stacking method to None, and number of EV-positive frames to 3.

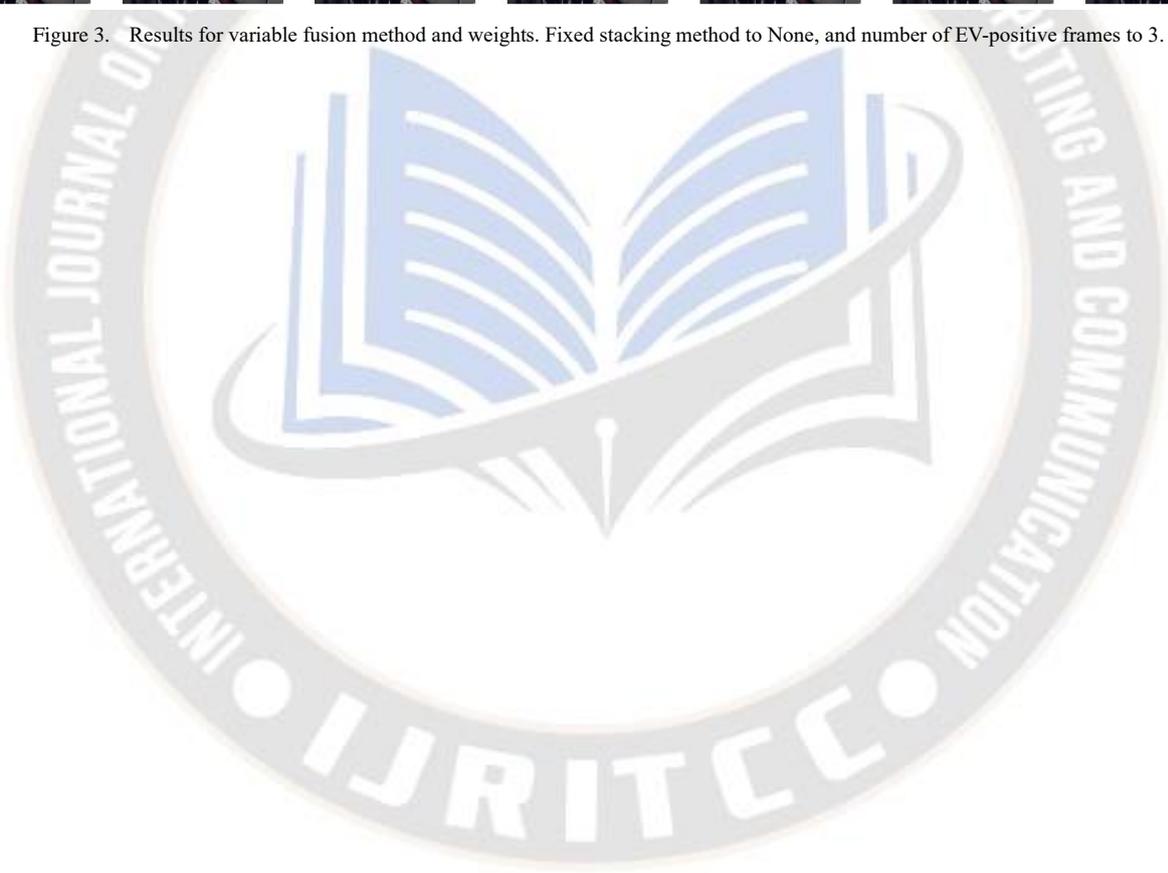

**3917**